\documentclass{article} 
\usepackage{iclr2023_conference,times}

\usepackage{amsmath,amsfonts,bm}

\def\eqref#1{equation~\ref{#1}}

\def\1{\bm{1}}

\def\vmu{{\bm{\mu}}}
\def\vsigma{{\bm{\sigma}}}

\def\vc{{\bm{c}}}

\def\vp{{\bm{p}}}

\def\vx{{\bm{x}}}

\def\vz{{\bm{z}}}

\def\mE{{\bm{E}}}

\def\mH{{\bm{H}}}
\def\mI{{\bm{I}}}

\def\mR{{\bm{R}}}

\def\mX{{\bm{X}}}

\DeclareMathAlphabet{\mathsfit}{\encodingdefault}{\sfdefault}{m}{sl}
\SetMathAlphabet{\mathsfit}{bold}{\encodingdefault}{\sfdefault}{bx}{n}
\newcommand{\tens}[1]{\bm{\mathsfit{#1}}}

\def\tC{{\tens{C}}}

\def\tX{{\tens{X}}}

\newcommand{\R}{\mathbb{R}}

\usepackage{hyperref}
\usepackage{url}
\usepackage{amsmath,amssymb,amsfonts,amsthm}
\usepackage{graphicx}
\usepackage{booktabs}    
\usepackage{soul}

\title{Multi-hypothesis 3D human pose estimation metrics favor miscalibrated distributions}

\author{Pawe\l\ A. Pierzchlewicz$^{1,2}$, R. James Cotton$^{3,4}$, Mohammad Bashiri$^{1,2}$, Fabian H. Sinz$^{1,2,5,6}$\\
$^1$Institute for Bioinformatics and Medical Informatics, Tübingen University, Tübingen, Germany\\
$^2$Department of Computer Science, Göttingen University, Göttingen, Germany\\
$^3$Shirley Ryan AbilityLab, Chicago, IL, USA\\
$^4$Department of Physical Medicine and Rehabilitation, Northwestern University, Evanston, IL, USA \\
$^5$Department of Neuroscience, Baylor College of Medicine, Houston, TX, USA\\
$^6$Center for Neuroscience and Artificial Intelligence, Baylor College of Medicine, Houston, TX, USA\\
\texttt{\{ppierzc,bashiri,sinz\}@cs.uni-goettingen.de}\\\texttt{rcotton@sralab.org}
}

\iclrpreprint
\begin{document}

\maketitle

\begin{abstract}
Due to depth ambiguities and occlusions, lifting 2D poses to 3D is a highly ill-posed problem.
Well-calibrated distributions of possible poses can make these ambiguities explicit and preserve the resulting uncertainty for downstream tasks. 
This study shows that previous attempts, which account for these ambiguities via multiple hypotheses generation, produce miscalibrated distributions.
We identify that miscalibration can be attributed to the use of sample-based metrics such as $\operatorname{minMPJPE}$.
In a series of simulations, we show that minimizing $\operatorname{minMPJPE}$, as commonly done, should converge to the correct mean prediction.
However, it fails to correctly capture the uncertainty, thus resulting in a miscalibrated distribution.
To mitigate this problem, we propose an accurate and well-calibrated model called Conditional Graph Normalizing Flow (cGNFs).
Our model is structured such that a single cGNF can estimate both conditional and marginal densities within the same model -- effectively solving a zero-shot density estimation problem.
We evaluate cGNF on the Human~3.6M dataset and show that cGNF provides a well-calibrated distribution estimate while being close to state-of-the-art in terms of overall $\operatorname{minMPJPE}$.
Furthermore, cGNF outperforms previous methods on occluded joints while it remains well-calibrated \footnote{Code and pretrained model weights are available at \url{https://github.com/sinzlab/cGNF}.}.
\end{abstract}

\section{Introduction}
The task of estimating the 3D human pose from 2D images is a classical problem in computer vision and has received significant attention over the years \citep{Agarwal2004-nc, Mori2006-gp, Bo2008-rg}.
With the advent of deep learning, various approaches have been applied to this problem with many of them achieving impressive results \citep{martinez_2017_3dbaseline, Pavlakos2016-zj, Pavlakos2018-jw, zhaoCVPR19semantic, Zou2021-nz}. 
However, the task of 3D pose estimation from 2D images is highly ill-posed: A single 2D joint can often be associated with multiple 3D positions, and due to occlusions, many joints can be entirely missing from the image.
While many previous studies still estimate one single solution for each image \citep{martinez_2017_3dbaseline, pavlakos2017harvesting, sun2017compositional, zhaoCVPR19semantic, Zhang2021DeepM3}, some attempts have been made to generate multiple hypotheses to account for these ambiguities \citep{Li_Hee_Lee_2019, Sharma_Varigonda_Bindal_Sharma_Jain_2019, Wehrbein_Rudolph_Rosenhahn_Wandt_2021, Oikarinen_Hannah_Kazerounian_2020, li2020weakly}.
Many of these approaches rely on estimating the conditional distribution of 3D poses given the 2D observation implicitly through sample-based methods.
Since direct likelihood estimation in sample-based methods is usually not feasible, different sample-based evaluation metrics have become popular.
As a result, the field's focus has been on the quality of individual samples with respect to the ground truth and not the quality of the probability distribution of 3D poses itself.

In this study, we show that common sample-based metrics in lifting, such as mean per joint position error,  encourage overconfident distributions rather than correct estimates of the true distribution.
As a result, they do not guarantee that the estimated density of 3D poses is a faithful representation of the underlying data distribution and its ambiguities.
As a consequence, their predicted uncertainty cannot be trusted in downstream decisions, which would be one of the key benefits of a probabilistic model.  

In a series of experiments, we show that a probabilistic lifting model trained with likelihood provides a higher quality estimated distribution.
First, we evaluate the distributions learned by minimizing $\operatorname{minMPJPE}$ instead of negative log-likelihood ($\operatorname{NLL}$) observing that, although the $\operatorname{minMPJPE}$ optimal distributions have a good mean they are not well-calibrated.
Next, we use the SimpleBaseline \citep{martinez_2017_3dbaseline} lifting model with a simple Gaussian noise model on Human3.6M to demonstrate that a model optimized for $\operatorname{NLL}$ is well-calibrated but underperforms on $\operatorname{minMPJPE}$. The same model optimized for $\operatorname{minMPJPE}$ performs well in that metric but turns out to be miscalibrated. 
To balance this trade-off, we propose an interpretable evaluation strategy that allows comparing sample-based methods, while retaining calibration.
Finally, we introduce a novel method to learn the distribution of 3D poses conditioned on the available 2D keypoint positions.
To that end, we propose a Conditional Graph Normalizing Flow (cGNF).
Unlike previous methods, cGNF does not require training a separate model for the prior and posterior.
Thus, our model does not require an adversarial loss term, as opposed to \citet{Wehrbein_Rudolph_Rosenhahn_Wandt_2021} and \citet{kolotouros2021}.
By evaluating the cGNF's performance on the Human 3.6M dataset \citep{h36m_pami},
we show that, in contrast to previous methods, our model is well calibrated while being close to state-of-the-art in terms of overall $\operatorname{minMPJPE}$, and that it significantly outperforms prior work in accuracy on occluded joints.

\section{Related work}
\paragraph{Lifting Models} Estimating the human 3D pose from a 2D image is an active research area \citep{Pavlakos2016-zj, martinez_2017_3dbaseline, zhaoCVPR19semantic, Wu2022-jb}.
An effective approach is to decouple 2D keypoint detection from 3D pose estimation \citep{martinez_2017_3dbaseline}. First, the 2D keypoints are estimated from the image using a 2D keypoint detector, then a lifting model uses just these keypoints to obtain a 3D pose estimate.
Since the task of estimating a 3D pose from 2D data is a highly ill-posed problem, approaches have been proposed to estimate multiple hypotheses \citep{Li_Hee_Lee_2019, Sharma_Varigonda_Bindal_Sharma_Jain_2019,  Oikarinen_Hannah_Kazerounian_2020, kolotouros2021, Li_Liu_Tang_Wang_Van_Gool_2021, Wehrbein_Rudolph_Rosenhahn_Wandt_2021}.
However, these approaches i) do not explicitly account for occluded or missing keypoints and ii) do not consider the calibration of the estimated densities.
\citet{Wehrbein_Rudolph_Rosenhahn_Wandt_2021} incorporate a Normalizing Flow~\citep{Tabak2000-ie} architecture to model the well-defined 3D to 2D projection and exploit the invertible nature of Normalizing Flows to obtain 2D to 3D estimates.
Albeit structured as a Normalizing Flow it is not trained as a probabilistic model.
Instead, the authors optimize the model by minimizing a set of cost functions. All in some form depend on the distance of hypotheses to the ground truth.
In addition, they utilize an adversarial loss to improve the quality of the hypotheses.
The proposed model achieves high performance on popular metrics in multi-hypothesis pose estimation, which are all sample-based distance measures rather than distribution-based metrics.
\citet{Sharma_Varigonda_Bindal_Sharma_Jain_2019} introduces a conditional variational autoencoder architecture with an ordinal ranking to disambiguate depth.
Similarly to \citet{Wehrbein_Rudolph_Rosenhahn_Wandt_2021}, the authors additionally optimize the poses on sample-based reconstruction metrics and report performance on sample-based metrics only.

\paragraph{Sample-Based Metrics in Pose Estimation} The most widely used metric in pose estimation is the mean per joint position error ($\operatorname{MPJPE}$) \citep{Wang2021-ps}.
It is defined as the mean Euclidean distance between the $K$ ground truth joint positions $\mX \in \R^{K\times3}$ and the predicted joint positions $\hat{\mX} \in \R^{K\times3}$.
Multi-hypothesis pose estimation considers $N$ hypotheses of positions $\hat{\tX} \in \R^{N \times K\times3}$ and adapts the error to consider the hypothesis closest to the ground truth \citep{Jahangiri2017-aq}.
\begin{equation*}
    \operatorname{minMPJPE}(\hat{\tX}, \mX) = \min_n \frac{1}{K} \sum_k^K \left|\left|\hat{\tX}_{n,k} - \mX_k\right|\right|_2
\end{equation*}
In this work, we refer to this minimum version of the $\operatorname{MPJPE}$ as $\operatorname{minMPJPE}$.
The percentage of correct keypoints ($\operatorname{PCK}$) \citep{Toshev2013-vi, Tompson2014-wh, Mehta2016-ji} is another widely accepted metric in pose estimation which measures the percentage of keypoints in a circle of 150mm around the ground truth in terms of $\operatorname{minMPJPE}$. Finally correct pose score ($\operatorname{CPS}$) proposed by \citet{Wandt2021Canonpose} considers a pose to be correct if all the keypoints are within a radius $r \in [0\textrm{ mm}, 300\textrm{ mm}]$ of the ground-truth in terms of $\operatorname{minMPJPE}$. $\operatorname{CPS}$ is defined as the area under the curve of percentage correct poses and $r$.

\paragraph{Calibration} Calibration is an important property of a probabilistic model.
It refers to the ability of a model to correctly reflect the uncertainty in the data.
Thus, the confidence of an event assigned by a well-calibrated model should be equal to the true probability of the event.
Humans have a natural cognitive intuition for probabilities \citep{Cosmides1996} and good confidence estimates can provide valuable information to the user, especially in safety-critical applications.
Therefore, density calibration has been an important topic in the machine learning community.
\citet{Guo2017-nv} show that calibration of densities has become especially important in the field of deep learning, where large models have been shown to be miscalibrated.
\citet{Brier1950} introduced the Brier score as a metric to measure the quality of a forecast.
It is defined as the expected squared difference between the predicted probability $\hat{\vp} \in \R^{N}$ and the true probability $\vp \in \R^{N}$ of $N$ samples.
\citet{Naeini2015} propose the expected calibration error ($\operatorname{ECE}$) metric which approximates the expectation of the absolute difference between the predicted probability and the true probability.
\begin{equation}
    \label{eq:ece}
    \operatorname{ECE} = \frac{1}{N} \sum_{n=1}^N \left| \hat{\vp}_n - \vp_n \right|
\end{equation}
The lower the $\operatorname{ECE}$ the better the calibration of the distribution.
A model which predicts the same probability for all samples has an $\operatorname{ECE}$ of 0.5, whereas a perfectly calibrated model has $\operatorname{ECE} = 0$.
\paragraph{Reliability diagrams} \citet{DeGroot1983} and \citet{Niculescu-Mizil2005} provide a visual representation of calibration. They display the calibration curve, which is a function of confidence against the true probability. If the calibration curve is an identity function then the model is perfectly calibrated.

\section{Observing Miscalibration}
\label{sec:observing_miscalibration}
    In this section, we demonstrate that the current state-of-the-art lifting models are not well calibrated.
    We consider two of the latest methods: \citet{Sharma_Varigonda_Bindal_Sharma_Jain_2019} and \citet{Wehrbein_Rudolph_Rosenhahn_Wandt_2021}.
    We compute the $\operatorname{ECE}$ for the two models and visualize their reliability diagrams (Fig~.\ref{fig:calibration}a).
    \subsection{Calibration for pose estimation}
\begin{figure}[t]
        \begin{center}
            \includegraphics[width=\linewidth]{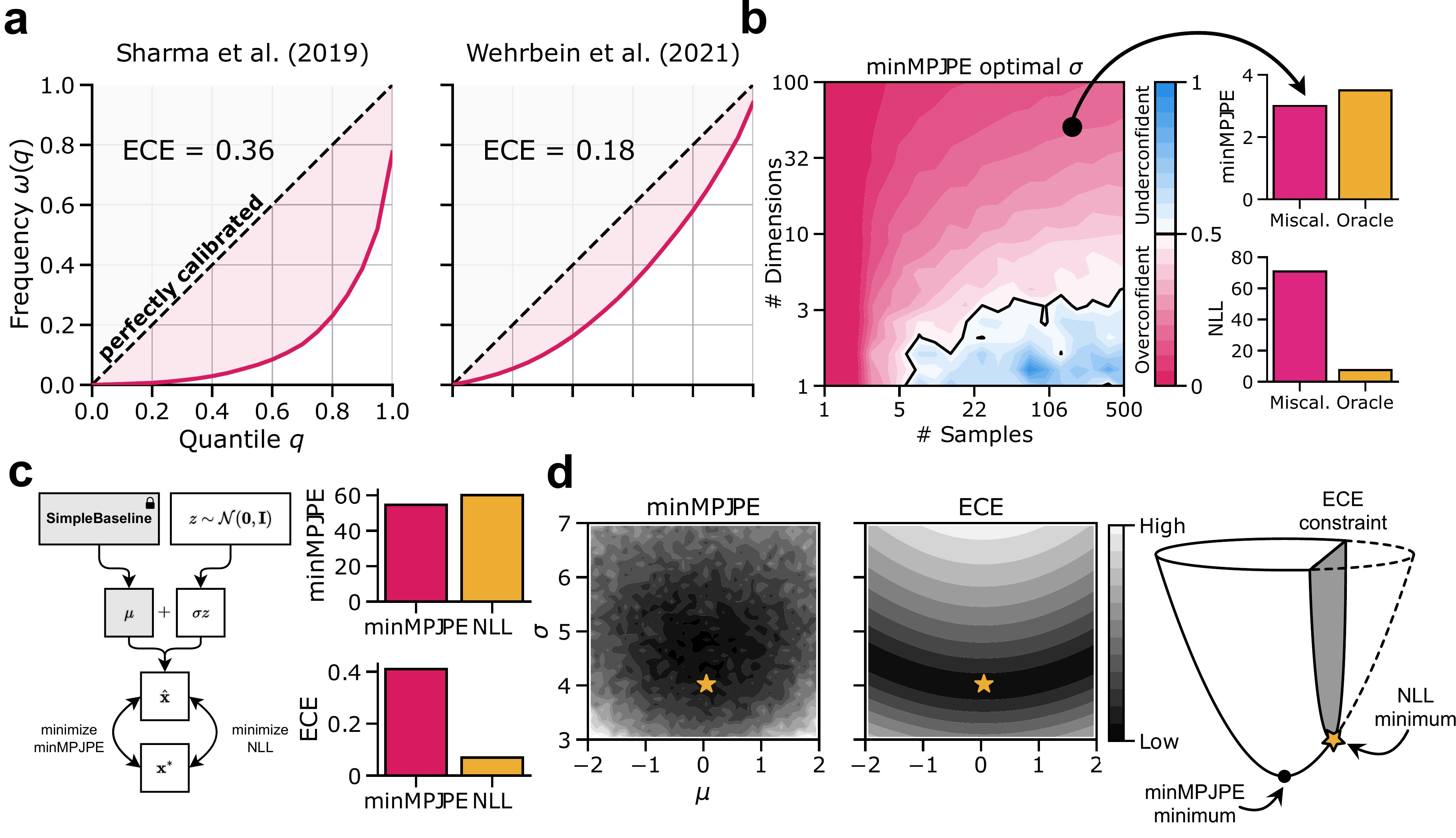}
        \end{center}
        \caption{
        \textbf{a}) Calibration curves of previous lifting models with the corresponding expected calibration error ($\operatorname{ECE}$) scores.
        \textbf{b}) Standard deviation $\sigma$ of a Gaussian distribution optimized to minimize $\operatorname{minMPJPE}$ for different number of samples and dimensions. The true $\sigma$ is 0.5 (black line), underconfident $\sigma > 0.5$ (blue), overconfident $\sigma < 0.5$ (pink). The human pose equivalent distribution (black point, 45 dimensions, 200 samples) compared to an oracle distribution (with true $\mu$ and $\sigma$) in terms of $\operatorname{minMPJPE}$ and $\operatorname{NLL}$.
        \textbf{c}) Gaussian noise model schematic to the left. The SimpleBaseline model weights are not trained. Right bar plots compare the performance on $\operatorname{minMPJPE}$ and $\operatorname{ECE}$ when optimizing for $\operatorname{minMPJPE}$ and $\operatorname{NLL}$.
        \textbf{d}) Loss landscapes of $\operatorname{minMPJPE}$ and $\operatorname{ECE}$ for a 1D Gaussian distribution with parameters $\sigma$ and $\mu$. Gold star represents the ground truth values of $\sigma^* = 4$ and $\mu^* = 0$. To the right is a schematic of the $\operatorname{ECE}$ constrained optimization.
        }
        \label{fig:calibration}
    \end{figure}

    We adapt the quantile calibration definition introduced in \citet{Song2019} for pose estimation problems.
    For $M$ 3D ground-truth poses with $K$ keypoints $\tX^* \in \R^{M\times K \times 3}$ we generate $N=200$ hypotheses $\hat{\tX} \in \R^{N \times M \times K \times 3}$ from the learned model $q(\tX \mid \tC)$ given the 2D pose $\tC \in \R^{M \times K \times 2}$.
    We compute the per-dimension median position $\tilde{\tX} \in \R^{M\times K\times3}$ of the hypotheses.
    Next, for each ground truth example $m$ and keypoint $k$ we compute the $L_2$ distance of each hypothesis $\hat{\tX}_{n,m,k}$ from the median $\bm{\varepsilon}_{n,m,k} = ||\hat{\tX}_{n,m,k} - \tilde{\tX}_{m, k}||_2$ to obtain a univariate distribution of errors.
    Using $\bm{\varepsilon}_{n,m,k}$ we obtain an empirical estimate of the cumulative distribution function $\Phi_m(\bm{\varepsilon})$.
    Given the distances $\bm{\varepsilon}^*_{m, k}$ of the ground truth $\tX^*_{m,k}$ from the median $\tilde{\tX}_{m, k}$ we compute the frequency $\omega_{k}(q)$ of $\bm{\varepsilon}^*_{:, k}$ falling into a particular quantile $q \in [0, 1]$:
    \begin{equation*}
        \omega_k(q) = \frac{1}{M} \sum^M_{m = 1} \1_\mathrm{\Phi_m(\bm{\varepsilon}^*_{m, k}) \le q},
    \end{equation*}
    Finally, we consider the median curve $\omega(q)$ across $K$ keypoints.
    An ideally calibrated model would result in $\omega(q) = q$.
    In this case, the error between the median estimate and ground truth would be consistent with the spread predicted by the inferred distribution.  
    With this formulation, we can compute the $\operatorname{ECE}$ according to \eqref{eq:ece}.
    We report the calibration curves $\omega(q)$ and $\operatorname{ECE}$s for each model in Fig.~\ref{fig:calibration}a.
  
    \subsection{Sample-based metrics promote miscalibration}
    \label{sec:simple_toy}
    Here, we show that sample-based metrics are a major component that contributes to miscalibration.
    In principle, $\operatorname{minMPJPE}$ could be a good surrogate metric for $\operatorname{NLL}$.
    However, as it became a common metric for selecting models it might become subject to Goodhart's Law \citep{Goodhart1975}  --  ``When a measure becomes a target, it ceases to be a good measure" \citep{Strathern1997-nr}.
    In the case of minimizing the mean $\operatorname{MPJPE}$ over hypotheses, the posterior distribution collapses onto the mean (sup.~\ref{app_seq:mpjpe}). Similarly, simulations indicate that $\operatorname{minMPJPE}$ converges to the correct mean, but it encourages miscalibration (Fig.~\ref{fig:calibration}b,d and \ref{app_sec:mpjpe_mean}).
    
    We illustrate this with a small toy example.
    Consider $M$ samples $\mX^* \in \R^{M\times D}$ from a $D$-dimensional Isotropic Normal distribution with mean $\vmu^{*} \in \mathbb{R}^D$ and variance $\vsigma^{*2} \in \mathbb{R}^D$ and an approximate isotropic Normal posterior distribution $q(\mX)$ with mean $\vmu \in \mathbb{R}^D$ and variance $\vsigma^2 \in \mathbb{R}^D$.
    We assume the ground truth mean to be known $\vmu = \vmu^{*}$ and only optimize the variance $\vsigma^2$ to minimize $\operatorname{minMPJPE}$ with $N$ hypotheses.
    We optimize $\vsigma^2$ for different numbers of dimensions $D$ and hypotheses $N$.
    Intuitively, for a small sampling budget drawing samples at the mean constitutes the least risk of generating a bad sample.
    With an increase in the number of hypotheses, increasing variance should gradually become beneficial, as the samples cover more of the volume.
    For a sufficiently large number of hypotheses, we can expect the variance to increase beyond the true variance, as the low probability samples can have sufficient representation.
    Increasing dimensions should have an inverse effect since the volume to be covered increases with each dimension.
    We observe these effects in the toy example (Fig.~\ref{fig:calibration}b).
    When we consider the case which corresponds to the 3D pose estimation problem ($D = 45$ and $N = 200$, black point in Fig. \ref{fig:calibration}b), we expect an overconfident distribution based on our toy example.
    This is also what we observe for the current state-of-the-art lifting models (Fig.~\ref{fig:calibration}a).
    Furthermore, we show that the $\operatorname{minMPJPE}$ optimal distribution outperforms the ground truth distribution in terms of $\operatorname{minMPJPE}$, but not in terms of negative log-likelihood (Fig.~\ref{fig:calibration}b).
    Together, the results imply that minimizing $\operatorname{minMPJPE}$, directly or by model selection, is expected to result in miscalibrated distributions and thus $\operatorname{minMPJPE}$ by itself is not sufficient to identify the best model.
    
    \subsection{Unconditional Gaussian noise baseline on human 3.6m}
    To verify the conclusions from the toy model in section~\ref{sec:simple_toy} we test the prediction with a simplified model on the Human3.6M dataset  \citep{IonescuSminchisescu11, h36m_pami} (see section \ref{sec:lifting} for more details about the dataset).
    We train an additive Gaussian noise model on top of the SimpleBaseline \citep{martinez_2017_3dbaseline} a well-established single-hypothesis model.
    We generate $N$ hypotheses $\hat{\tX} \in \R^{N\times M \times K \times 3}$ of poses with $K$ keypoints for $M$ observations $\tC \in \R^{M \times K \times 2}$ according to:
    \begin{equation*}
        \hat{\tX}_{n, m} = \operatorname{SimpleBaseline}(\tC_m) + \vsigma \vz_n
    \end{equation*}
    where $\operatorname{SimpleBaseline}(\tC_m)$ estimates the mean of the noise, $\vsigma$ is the standard deviation parameter scaling the standard normal samples $\vz \sim \mathcal{N}(\vz; \mathbf{0}, \mI)$ (Fig.~\ref{fig:calibration}c).
    It is important to note that we do not condition $\vsigma$ on the 2D observation $\tC_m$, i.e. the same noise model is used for every input.
    We test two optimization setups: 1) minimizing $\operatorname{minMPJPE}$ and 2) maximizing likelihood.
    Based on the predictions from the toy problem (sec. \ref{sec:observing_miscalibration}), we expect the $\operatorname{minMPJPE}$ model to be overconfident and outperform the $\operatorname{NLL}$ model on the $\operatorname{minMPJPE}$, but the $\operatorname{NLL}$ model to be better calibrated. 
    This is exactly what we observe (Fig.~\ref{fig:calibration}c).
    Furthermore, each of these models achieves $\operatorname{minMPJPE}$ performance in a range similar to \citet{Sharma_Varigonda_Bindal_Sharma_Jain_2019} and \citet{Wehrbein_Rudolph_Rosenhahn_Wandt_2021} (Table~\ref{tab:h36m_results}).
    
    \subsection{Evaluating sample-based methods}
    Given that $\operatorname{minMPJPE}$ is not sufficient to fully evaluate multi-hypothesis methods, we propose a strategy that remains interpretable and promotes calibrated distributions.
    Consider the landscapes of $\operatorname{minMPJPE}$ and $\operatorname{ECE}$ with respect to the mean and variance of an approximate distribution (Fig. \ref{fig:calibration}d). Simulations indicate that optimizing $\operatorname{minMPJPE}$ identifies the correct mean $\mu$ (sup.~\ref{app_sec:mpjpe_mean}), but not  the correct $\sigma$. 
    $\operatorname{ECE}$, however, is minimized by a manifold of $\mu, \sigma$ values and converges to a good standard deviation for each mean.
    We thus hypothesize that a likelihood-optimal distribution can be approximated when $\operatorname{minMPJPE}$ is minimized on the $\operatorname{ECE}$-optimal manifold.
    Hence, $\operatorname{minMPJPE}$ reflects only one dimension of the performance metric.

\section{Conditional graph normalizing flow}
    The human pose has a natural graph structure, where the nodes represent joints and edges represent bone connections between joints.
    In this section, we introduce a method that utilizes the graph representation of the human pose for 3D estimation via likelihood maximization.
    We propose to learn the conditional distribution $p(\vx \mid \vc)$ of the 3D pose $\vx$ given the 2D pose $\vc$ using conditional graph normalizing flows (cGNF).
    
\begin{figure}[t!]
\centering
\includegraphics[width=\linewidth]{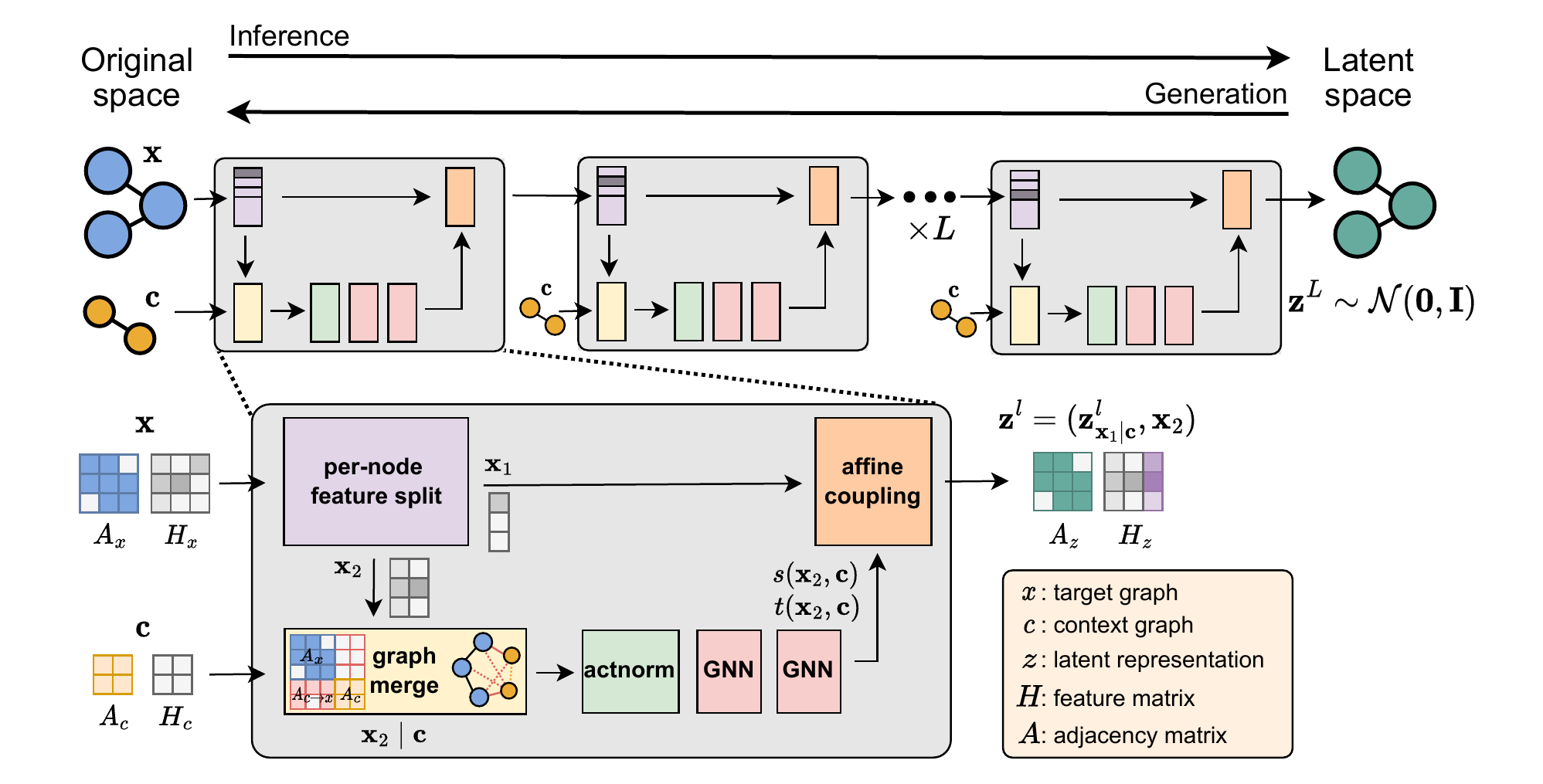}
\caption{
 A schematic of the cGNF. Target variables $\vx$ are represented by a graph with the feature matrix $H_{x}$ and the adjacency matrix $A_x$. The context variables are represented by a context graph $\vc$ with the feature matrix $H_c$ and adjacency matrix $A_x$. In the inference path the target graph $\vx$ is transformed into a latent space $\vz$ which follows a standard normal distribution. The transformation is achieved through $L$ transformation blocks.
}

\label{fig:fig1}
\end{figure}

    We define a target graph $\vx = (\mH^x, \mE^x)$ of 3D poses and a context graph $\vc = (\mH^c, \mE^c)$ of 2D detections.
    $\mH^x \in \mathbb{R}^{n\times D_x}$ and $\mE^x$ are the edges between the nodes of the target graph and $\mH^c \in \mathbb{R}^{m\times D_c}$ and $\mE^c$ are the edges between the nodes of the context graph.
    In the case that an observation is not present, the corresponding node is removed from $\vc$.
    The model is built of $L$ transformation blocks, each of which consists of a per-node feature split step, a graph merging step, an $\operatorname{actnorm}$ \citep{Kingma_Dhariwal_2018} and two graph neural network layers \citep{Gori2005} (Fig.~\ref{fig:fig1}).
    These elements construct an affine coupling layer \citep{Dinh_Sohl-Dickstein_Bengio_2016}, which is then followed by a permutation layer.
    The transformation blocks are only applied to the target graph, while the context graph is passed through unchanged.
    
    \paragraph{Per-Node Feature Split Step} splits the target node features $\mH_x$ into two parts, $\mH_x^{(1)}$ and $\mH_x^{(2)}$ across the feature dimension.
    We incorporate a leave-one-out strategy for splitting the features.
    The $i$th feature dimension is propagated directly to the affine coupling layer and the remaining dimensions are passed to the graph neural network layers.
    In the next block, the next $i$th dimension is used.

    \paragraph{Graph Merging}
    When utilizing conditional normalizing flows \cite{Winkler_Worrall_Hoogeboom_Welling_2019} on graph-structured data, a key challenge is incorporating the context graph in the transformation.
    We propose to merge the context graph $\mathbf c$ with the target graph $\vx$ into a heterogeneous graph $\vx\mid\vc$.
    The context graph $\mathbf c$ forms directed edges from nodes in $\vc$ to nodes in $\vx$ as defined by $\mR^{c\rightarrow x}$, the relations matrix.
    $\mR^{c\rightarrow x}_{i,j} = 1$ indicates that node $i$ in the context graph forms an edge with node $j$ in the target graph $\vx$ (Fig.~\ref{fig:fig1}).
    
    \paragraph{Graph Neural Network Layers}
    We define the graph neural network layers as relational graph convolutions (R-GCNs) \citep{schlichtkrull2017modeling}.
    In the message passing step, the message received by node $v$ from the neighboring nodes is defined as
    \begin{equation*}
        \mathbf{m}^{(v)}_{t+1}=\sum_{u \in \mathcal{N}^{c\rightarrow x}(v)}\psi_{c\rightarrow x}\left(\mathbf{h}_{\vc}^{(v)}, e^{(u, v)}\right) + \sum_{r \in \mathcal{R}} \sum_{u\in\mathcal{N}^r(v)} \psi_r\left(\mathbf{h}^{(u)}_t, e^{(u, v)}\right)\label{eq:cgnf_message}
    \end{equation*}
    where $\psi_r : \mathbb{R}^{D_n} \mapsto \mathbb{R}^{D_h}$ and $\psi_{c\rightarrow x} : \mathbb{R}^{D_c} \mapsto \mathbb{R}^{D_h}$, with $D_h$ as the number of latent dimensions.
    $\psi_{c\rightarrow x}$ should be flexible enough to allow the network to learn to distinguish between missing observations and zero observations i.e. $\psi_{c\rightarrow x}\left(\mathbf{0}, e^{(u, v)}\right) \ne \mathbf{0}$.
    The $\operatorname{Update}$ step is defined by the mapping $g: \mathbb{R}^{D_h} \mapsto \mathbb{R}^{D_o}$ which maps the latent space to the output dimension of size $D_o$.
    We implement the $\operatorname{Update}$ step as a single fully connected linear layer.

    \paragraph{Affine Coupling Layer}
    Similarly to \citet{Liu_Kumar_Ba_Kiros_Swersky_2019} the output of the GNN layers models the scale $\mathbf{s}(\vx_2, \vc)$ and translation $\mathbf{t}(\vx_2, \vc)$ functions.
    The scale and translation functions are then applied to the unchanged split $\vx_1$ to produce the transformed graph $\vz_1^{l}$.
    \begin{align*}
        \vz_1^{l} &= \vx_1 \odot \exp\left(\mathbf{s}(\vx_2, \vc)\right) + \mathbf t(\vx_2, \vc)\\
        \vz_2^{l} &= \vx_2
    \end{align*}
    The $\vx_2$ is copied to $\vz_2^{l}$ unchanged.
    The $\vz_1^{l}$ and $\vz_2^{l}$ are then concatenated to form the transformed graph $\vz^{l}$, which is passed to the next transformation block.
    
    \paragraph{Loss}
    The standard optimization procedure for normalizing flows is to maximize the log probability of the observed data $\vx$ obtained through the inverse path ($\vx \rightarrow \vz$) (Fig.~\ref{fig:fig1}).
    Assuming $\vx$ are i.i.d. the task of the flow is to model $p\left(\vx \mid \vc\right) = \prod_i^N p\left(\vx_i \mid \vc_i\right)$
    where $\vx_i$ are the 3D poses and $\vc_i$ are the corresponding 2D observations.
    We thus define the loss as the negative log probability of pairs of observations $\vx$ and $\vc$.
    \begin{equation*}
        \mathcal{L}_{post.} = -\ln q_0(f(\vx, \mathbf c)) + \sum^K_{k=1} \ln \left| \textrm{det} \nabla_{z_{k-1}} f_k(\vz_{k-1}, \vc)\right|
    \end{equation*}
    where $q_0 \sim \mathcal{N}(\vz; \mathbf{0}, \mI)$ is the source distribution.
    We augment the training data by randomly removing context variables to simulate new observations with missing keypoints in $\vc$.
    The augmented observations contain $20\%, 40\%, 60\%$ or $80\%$ of all observable keypoints.
    For all 3D poses, we additionally compute the prior loss, which expresses the likelihood of a pose given that no 2D keypoints were observed.
    \begin{equation*}
        \mathcal{L}_{prior} = -\ln q_0(f(\vx, \varnothing)) + \sum^K_{k=1} \ln\left| \textrm{det} \nabla_{z_{k-1}} f_k(\vz_{k-1}, \varnothing)\right|
    \end{equation*}
    Our overall loss function is thus the sum of the two partial losses. 
    \begin{equation}
    \label{eq:loss}
        \mathcal{L} = \frac{1}{2}\Big(\mathcal{L}_{prior} + \mathcal{L}_{post}\Big)
    \end{equation}
    The proposed training strategy and architecture formulate pose estimation as a zero-shot density estimation problem.
    The cGNF model is trained on a subset of possible observations and is required to evaluate previously unseen conditional densities.
    We explore these zero-shot capabilities in the appendix (sup~\ref{sec:zero_shot}).
    
    \paragraph{Root Node}
    3D poses are relative to a root node (usually the pelvis).
    Hence, the root node's position is deterministic.
    We therefore remove the root node and corresponding edges from the target graph $\vx$ and represent it as a \textit{root} node-type $\mathbf{r}$, which has features $\mH^\mathbf{r} \in \mathbb{R}^3$ and a message generation function $\psi_r$ which is a fully connected neural network with 100 units.

    \paragraph{Graph Symmetries}
    The human pose graph has symmetries, e.g. the left and right limbs are mirrored.
    We impose a hierarchical structure on the nodes of the target graph $\vx$.
    A node may have a parent and a child, for example, the elbow node is the child of the shoulder node and the parent of the wrist node.
    Messages passed from the parent to the child are \textit{forward} messages generated by $\psi_{x\rightarrow x}$ and messages from the child to the parent are \textit{backward} messages generated by $\psi_{x\leftarrow x}$.

    \paragraph{Occlusion Representation}
    We use 2D keypoint positions published by \citet{Wehrbein_Rudolph_Rosenhahn_Wandt_2021} estimated using the HRNet model \citep{sun2019deep} and the provided Gaussian distribution fits for evaluating occluded keypoints.
    If a keypoint is classified as occluded (2D detection $\sigma > 5$px) its corresponding node is removed from the context graph.
    To adjust for the differences between the pose definitions used by HRNet and H36M we employ an embedding network using the SageConv architecture \citep{hamilton_ying_leskovec} with a learnable adjacency matrix.
    The embedding network transforms the observed 2D keypoints into a 10-dimensional embedding vector for each of the keypoints. Additional implementation details of the architecture are given in the appendix (\ref{sec:app_arch_dets}).

\section{Lifting Human3.6M}
\label{sec:lifting}

\paragraph{Data} We use the Human3.6M Dataset (H36M) on the academic use only license \citep{IonescuSminchisescu11, h36m_pami} which is the largest dataset for 3D human pose estimation.
It consists of tuples of 2D images, 2D poses, and 3D poses for 7 professional actors performing 15 different activities captured with 4 cameras.
Accurate 3D positions are obtained from 10 motion capture cameras and markers placed on the subjects.
For evaluation, we additionally use the Human 3.6M Ambiguous (H36MA) dataset introduced by \citet{Wehrbein_Rudolph_Rosenhahn_Wandt_2021}.
H36MA is a subset of the H36M dataset containing only ambiguous poses from subjects 9 and 11.
A pose is defined as ambiguous when the 2D keypoint detector is highly uncertain about at least one of the keypoints.

\paragraph{Training}
We train the model on subjects 1, 5, 6, 7, and 8 on every 4th frame.
We reduce the learning rate on plateau with an initial learning rate of 0.001 and patience of 10 steps reducing the learning rate by a factor of 10.
Training is stopped after the 3rd decrease in the learning rate or 200 epochs.
The model was trained on a single Nvidia Tesla V100 GPU, for about 6 days.

\begin{figure}[t!]
\centering
\includegraphics[width=\linewidth]{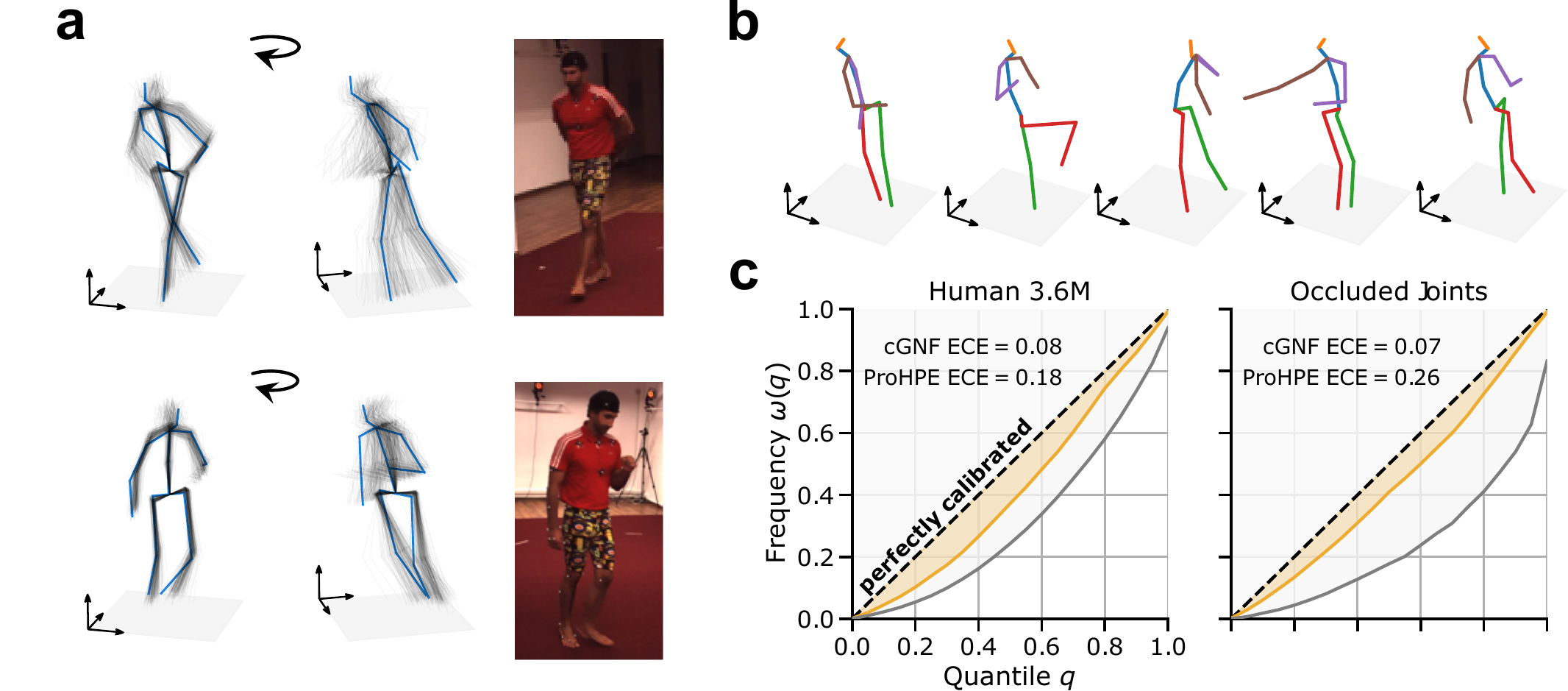}
\caption{\textbf{a)} Hypotheses generated by the cGNF (gray) vs the ground truth pose (blue). Original image is shown to the right.
\textbf{b)} Example of samples from the prior learned by the cGNF.
\textbf{c)} Calibration of the conditional density. Comparison of the frequency that the distance of the ground truth from the median pose is within a given quantile. Median calibration curves for our model (cGNF, orange) and \citet{Wehrbein_Rudolph_Rosenhahn_Wandt_2021} (ProHPE, gray). Left shows calibration curves for the whole Human 3.6M test set and right for only the occluded joints.
}
\label{fig:fig3}
\end{figure}

\begin{table}[h!]
  \def\arraystretch{1.25}
  \caption{Comparison of the cGNF model to state-of-the-art methods for multi-hypothesis pose estimation using expected calibration error ($\operatorname{ECE}$) and minimum mean per joint position error ($\operatorname{minMPJPE}$) between the ground truth 3D pose and $N$ hypotheses.
  Best model row is printed in bold font.
  Reporting the mean across the outcomes of 3 different seeds and the standard deviation (SD). For $\operatorname{ECE}$ the SD is smaller than $0.001$ in all cases. Thus, we do not report the SD value for $\operatorname{ECE}$ in this table.
  For all the metrics lower is better. We underlined the results that we did not compute but instead used the originally reported value.\\}
  \label{tab:h36m_results}
  \centering
  \scriptsize
\begin{tabular}{l|lll|ll|ll}
\toprule
Method                                        & H36M \tiny{(mm)}           & H36MA \tiny{(mm)}          & $\operatorname{ECE}$                & Occluded \tiny{(mm)}                      & $\operatorname{ECE}$                & $N$               & \# Params \\
\midrule
\citet{martinez_2017_3dbaseline} & \underline{62.9} & - & - &- &- &1&4,288,557\\
\midrule
\citet{Li_Hee_Lee_2019}                            & \underline{52.7}                      & \underline{81.1}                      & -                                & -                                       & -                               & 5                        & 4,498,682 \\
\citet{Sharma_Varigonda_Bindal_Sharma_Jain_2019} & \underline{46.7}                       & \underline{78.3}                       & 0.36 \tiny{(72\%)}                       & -                                       & -                               & 200                      & 9,100,080 \\
\citet{Wehrbein_Rudolph_Rosenhahn_Wandt_2021}     & \underline{\textbf{44.3}}                      & \underline{\textbf{71.0}}                    & 0.18 \tiny{(36\%)}                       & 51.1 \tiny{$\pm$ 0.13}                     & 0.26 \tiny{(52\%)}                      & 200                      & 2,157,176 \\ 
\midrule
Gaussian ($\operatorname{minMPJPE}$)       & 54.8 \tiny{$\pm$ 0.002}  & -       & 0.42 \tiny{(82\%)}  & -                  & -          & 200 & 4,288,572 \\
Gaussian ($\operatorname{NLL}$)       & 60.1 \tiny{$\pm$ 0.002}  & -       & 0.07 \tiny{(14\%)}  & -                  & -          & 200 & 4,288,572 \\
\midrule
cGNF                    & 57.5 \tiny{$\pm$ 0.06} & 87.3 \tiny{$\pm$ 0.13} & \textbf{0.08} \tiny{(16\%)}  & 47.0 \tiny{$\pm$ 0.18} & 0.07 \tiny{(14\%)}  & 200 & 852,546   \\
cGNF w $\mathcal{L}_{sample}$ & 53.0 \tiny{$\pm$ 0.06} & 79.3 \tiny{$\pm$ 0.05} & \textbf{0.08} \tiny{(16\%)}  & 41.8 \tiny{$\pm$ 0.04} & \textbf{0.03} \tiny{(6\%)}  \ \  & 200 & 852,546 \\
cGNF xlarge w $\mathcal{L}_{sample}$                                  & 48.5  \tiny{$\pm$ 0.02}                     & 72.6 \tiny{$\pm$ 0.09}                      & 0.23 \tiny{(46\%)}                        & \textbf{39.9} \tiny{$\pm$ 0.05}                                & 0.07 \tiny{(14\%)}                      & 200                      & 8,318,741 \\
\bottomrule
\end{tabular}
\end{table}

\paragraph{Evaluation}
We evaluate the model on every 64th frame of subjects 9 and 11 and the H36MA subset.
We compare our model's performance to prior work on $\operatorname{minMPJPE}$ and $\operatorname{ECE}$ using 200 samples (Table~\ref{tab:h36m_results}).
As expected from the observations made in section \ref{sec:observing_miscalibration}, our method underperforms on $\operatorname{minMPJPE}$ but significantly outperforms on $\operatorname{ECE}$ (Fig.~\ref{fig:fig3}c).
Samples from the posterior and prior are shown in figure \ref{fig:fig3}a and b.
Additional examples are included in the appendix (posterior samples Fig.~\ref{fig:supp_posterior}; prior samples Fig.~\ref{fig:prior_samples}).

\paragraph{Performance on individual occluded joints}
The poses contained in H36MA are not only occluded but also generally more difficult than the average pose in H36M.
Therefore, we propose to evaluate the performance on solely the occluded joints instead of the whole poses.
We report these errors in table \ref{tab:h36m_results} (Occluded), where we show that our method outperforms the competing methods by a significant margin on both $\operatorname{minMPJPE}$ and $\operatorname{ECE}$.
Thus, this shows that our model is able to learn a posterior distribution that is more calibrated than previous methods and is able to outperform prior methods on $\operatorname{minMPJPE}$ for the occluded joints.

\paragraph{Improving minMPJPE performance and the effect on calibration}
We can incorporate a couple of additional steps to improve the $\operatorname{minMPJPE}$ performance.
We introduce an additional loss term 
\begin{equation*}
    \mathcal{L}_{sample} = \operatorname{MPJPE}\left(\vx^*, f^{-1}(\mathbf{0}, c)\right)
\end{equation*} 
that encourages the model to predict the ground truth pose.
The sample-based loss term is added to the vanilla loss (\eqref{eq:loss}) with a scaling coefficient $\lambda_{sample} = 0.1$. 
Analogously to \citet{kolotouros2021} we sample a pose from the mode of the source distribution and minimize the $\operatorname{minMPJPE}$ between the sampled pose and the ground truth pose.
This additional loss term is shown to improve the $\operatorname{minMPJPE}$ performance. 
At the original model capacity, the $\operatorname{minMPJPE}$ and calibration performance show improvement.
However, while the model performance on $\operatorname{minMPJPE}$ increases further with model capacity, calibration decreases significantly.
We compare the performances in table~\ref{tab:h36m_results}.
Additional model capacity evaluations are made in sup.~\ref{sec:app_model_scale}.

\section{Conclusion}
    In this study, we explored the problem of miscalibration in multi-hypothesis 3D pose estimation.
    Obtaining calibrated density estimates is important for safety-critical applications, such as healthcare or autonomous driving.
    Here we provide evidence that a focus on sample-based metrics for multi-hypothesis 3D pose estimation (e.g. $\operatorname{minMPJPE}$) can lead to miscalibrated distributions.
    We propose a flexible model which can be trained to minimize the negative log-likelihood loss and
    show that, unlike previous methods, our model can learn a well-calibrated posterior distribution at a small loss of overall accuracy.
    However, in particularly ambiguous situations, i.e. for the occluded joints, we show that our model outperforms the state-of-the-art on $\operatorname{minMPJPE}$ while maintaining a well-calibrated distribution.
    We believe that our findings will be useful for future work in identifying and mitigating miscalibration in multi-hypothesis pose estimation and will lead to more robust and safer applications of multi-hypothesis pose estimation.

\subsubsection*{Acknowledgments}
We thank Alexander Ecker, Pavithra Elumalai, Arne Nix, Suhas Shrinivasan and Konstantin Willeke for their helpful feedback and discussions.

Funded by the German Federal Ministry for Economic Affairs and Climate Action (FKZ ZF4076506AW9). This work was supported by the Carl-Zeiss-Stiftung (FS). The authors thank the International Max Planck Research School for Intelligent Systems (IMPRS-IS)
for supporting MB.

\bibliography{main}
\bibliographystyle{iclr2023_conference}

\appendix
\renewcommand{\figurename}{Supplementary Figure}
\renewcommand{\tablename}{Supplementary Table}
\section{Metrics}
\subsection{Mean Per Joint Position Error}
\label{app_seq:mpjpe}
A popular optimization metric is the $\operatorname{MPJPE}$.
While this metric is especially popular in single-pose estimation methods, it has also been used in various forms in multi-hypothesis methods.
Optimizing this metric causes the distribution of poses to be overconfident.
We show this for a simple one-dimensional distribution, the generalization to the multi-dimensional case is straightforward. 
Given samples $x\sim p(x\vert c)$ from a data distribution given a particular context $c$, such as keypoints from a image,  consider an approximate distribution $q(\hat{x}\vert c)$ supposed to reflect the uncertainty about $x|c$. 

This below objective is equivalent to the mean position error for a single joint. Note that $x$ and $\hat x$ are conditionally independent given $c$, i.e. $x \bot \hat x\vert c$. The objective can then be expanded as follows:
\begin{align*}
    \mathcal{L} &= \mathbb{E}_{x\sim p(x|c), \hat{x}\sim q(\hat x|c), c}\left[(x - \hat{x})^2\right]\\
    &= \mathbb{E}_{c}\left[\mathbb{E}_{x, \hat{x} \mid c}\left[(x - \mu_c + \mu_c - \hat{x})^2\right]\right]\\
    &= \mathbb{E}_c\left[\underbrace{\operatorname{Var}[x \mid c]}_{\textrm{indep. of } q} - 2\mathbb{E}_{x,\hat{x}\mid c}\left[(x - \mu_c)(\hat{x} - \mu_c)\right] + \mathbb{E}_{\hat{x}\mid c}\left[(\hat{x} - \mu_c)^2\right]\right]\\
    &= \mathrm{const.} - 2\mathbb{E}_c\left[\underbrace{\mathbb{E}_{x\mid c}\left[(x - \mu_c)\right]}_{=0}\mathbb{E}_{\hat{x}\mid c}\left[(\hat{x} - \mu_c)\right] + \mathbb{E}_{\hat{x}\mid c}\left[(\hat{x} - \mu_c)^2\right]\right]\\
    &= \mathrm{const.} + \mathbb{E}_{c}\left[\mathbb{E}_{\hat{x} \mid c}\left[(\hat{x} - \mu_c)^2\right]\right] \ge 0
\end{align*}
The expectation in the final line is non-negative and can be minimized by $q(\hat x|c) = \delta(\hat x - \mu_c)$, i.e. setting $\hat x = \mu_c$ and shrinking the variance to zero. This means that $q$ would be extremely overconfident. 

\subsection{$\operatorname{minMPJPE}$ converges to the correct mean}
\label{app_sec:mpjpe_mean}
Consider 1D samples $x^*$ from a data distribution $p(x)$ and an approximate Gaussian distribution $q(x)$ with parameters $\mu$ and $\sigma$.
We sample $N$ hypotheses from $q(x)$ and minimize the $\operatorname{minMPJPE}$ objective:
\begin{align*}
    \operatorname{minMPJPE} &= \mathbb{E}_{q(z)}\left[\mathbb{E}_{p(x)}\left[\min_i (x^* - \mu - \sigma z_i)^2\right]\right]
\end{align*}
Consider $z_j^*$ as the $z_i$ sample which minimizes the expression for the $j$-th data sample $x^*_j$.
\begin{align*}
    \operatorname{minMPJPE} &= \mathbb{E}_{q(z)}\left[\mathbb{E}_{p(x)}\left[(x^* - \mu - \sigma z^*_j)^2\right]\right]
\end{align*}
Thus the derivative can be computed to be
\begin{align*}
    \frac{\partial}{\partial \mu}\operatorname{minMPJPE} &= -2\mathbb{E}_{q(z)}\left[\mathbb{E}_{p(x)}\left[x^* - \mu - \sigma z^*_j\right]\right] = 0\\
    &= \mathbb{E}_{p(x)}\left[x^*\right] - \mu - \mathbb{E}_{q(z)}\left[z^*_j\right]
\end{align*}
Simulations indicate that $\mathbb{E}_{q(z)}\left[z^*_j\right]$ can be approximated by a sigmoid function
\begin{equation*}
    \mathbb{E}_{q(z)}\left[z^*_j\right] = S\left(\mathbb{E}_{p(x)}\left[x^*\right] - \mu\right) \cdot C(\sigma, N)
\end{equation*}
where $C(\sigma, N)$ is a scalar scaling value dependent on $\sigma$ and the number of hypotheses.
Thus the root of the derivative can be computed to be:
\begin{equation*}
    \mu = \mathbb{E}_{p(x)}\left[x^*\right] 
\end{equation*}

\section{Conditional Graph Normalizing Flow}
\subsection{Architecture Details}
\label{sec:app_arch_dets}
The cGNF model consists of 10 flow layers.
Each flow layer $f_k$ consists of two GNN layers each performing one message-passing step each as defined in eq. \eqref{eq:cgnf_message}.
In the first GNN layer each message generation function $\psi_r^{(1)}$ is a single layer fully-connected neural network with 100 units and a ReLU activation \citep{agarap2018deep}. All the messages to a node are summed together resulting in the output of the $\operatorname{Message}$ as in eq. \eqref{eq:cgnf_message}. Then the $\operatorname{Update}$ step takes the message output as its input to a single-layer fully connected neural network with 100 units and linear activation.
The context $\vc$ is transformed via $\psi_c$ to 100 dimensions and passed to the next GNN layer.
In the next GNN layer, the message generation functions $\psi_r^{(2)}$ are single layer fully connected neural networks with 100 units and ReLU activation,
The $\operatorname{Update}$ is a neural network layer with 3 output units.
In the next flow layer of the original context graph $\vc$ is used and not the transformed context.

\subsection{Zero-shot density estimation}
\label{sec:zero_shot}
We evaluate cGNF's zero-shot capability to estimate a previously unseen conditional density.
We simulated 50 different triple pendulums with initial velocities sampled from a normal distribution $v \sim \mathcal{N}(0, 10)$ for 25 timesteps each.
Each pendulum was constructed from 4 nodes connected in a chain. The zeroth node was fixed and the remaining $x_1$, $x_2$ and $x_3$ were freely moving.
The nodes were observed with $c_i = x_i + \varepsilon$ with $\varepsilon \sim \mathcal{N}(0, 5 \cdot 10^{-2})$.
On this dataset, we trained 3 models. I. A CNF trained to estimate the density when all positions are observed $p(x \mid c_1, c_2, c_3)$ II. A CNF trained on a density where only one node is observed $p(x \mid c_1)$ III. A cGNF trained on the densities where at most 2 nodes are observed i.e. the cGNF never sees examples of $p(x \mid c_1, c_2, c_3)$.

\begin{figure}[t]
\centering
\includegraphics[width=\linewidth]{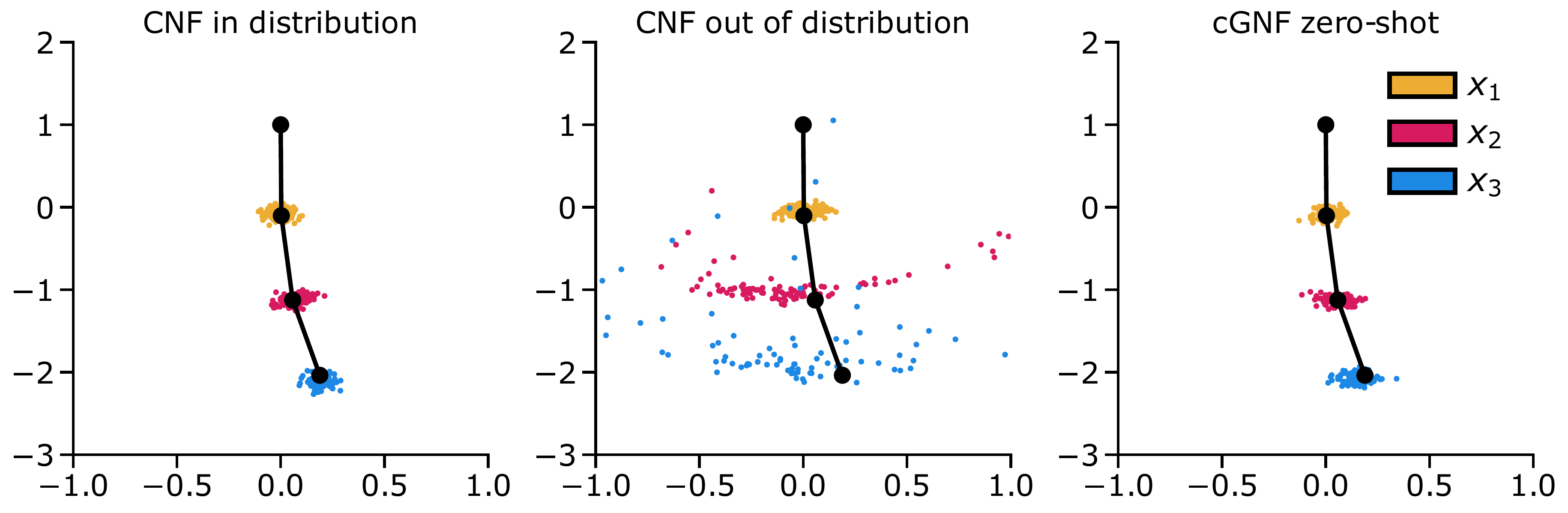}
\caption{Zero-shot capabilities of the cGNF model. Density estimates for 3 models: 1) CNF trained on $p(x | c_1, c_2, c_3)$ 2) CNF trained on $p(x | c_1)$ 3) cGNF which has never seen $p(x | c_1, c_2, c_3)$. The black points represent the true positions of the triple pendulum, and the orange, magenta and cyan represent samples for each node $x_1$, $x_2$, $x_3$ respectively.}
\label{fig:fig4}
\end{figure}

To test zero-shot capabilities we compare the performances of these 3 models on $p(x \mid c_1, c_2, c_3)$. Model I (CNF) is used as reference for estimating this distribution when $p(x \mid c_1, c_2, c_3)$ is in distribution. Model II (CNF) is used to reference a model which cannot zero-shot estimate densities as it is out of distribution. Model III (cGNF) shows that our model can zero-shot estimate a previously unseen conditional density (Fig.~\ref{fig:fig4}).

\begin{figure}[h]
\centering
\includegraphics[width=\linewidth]{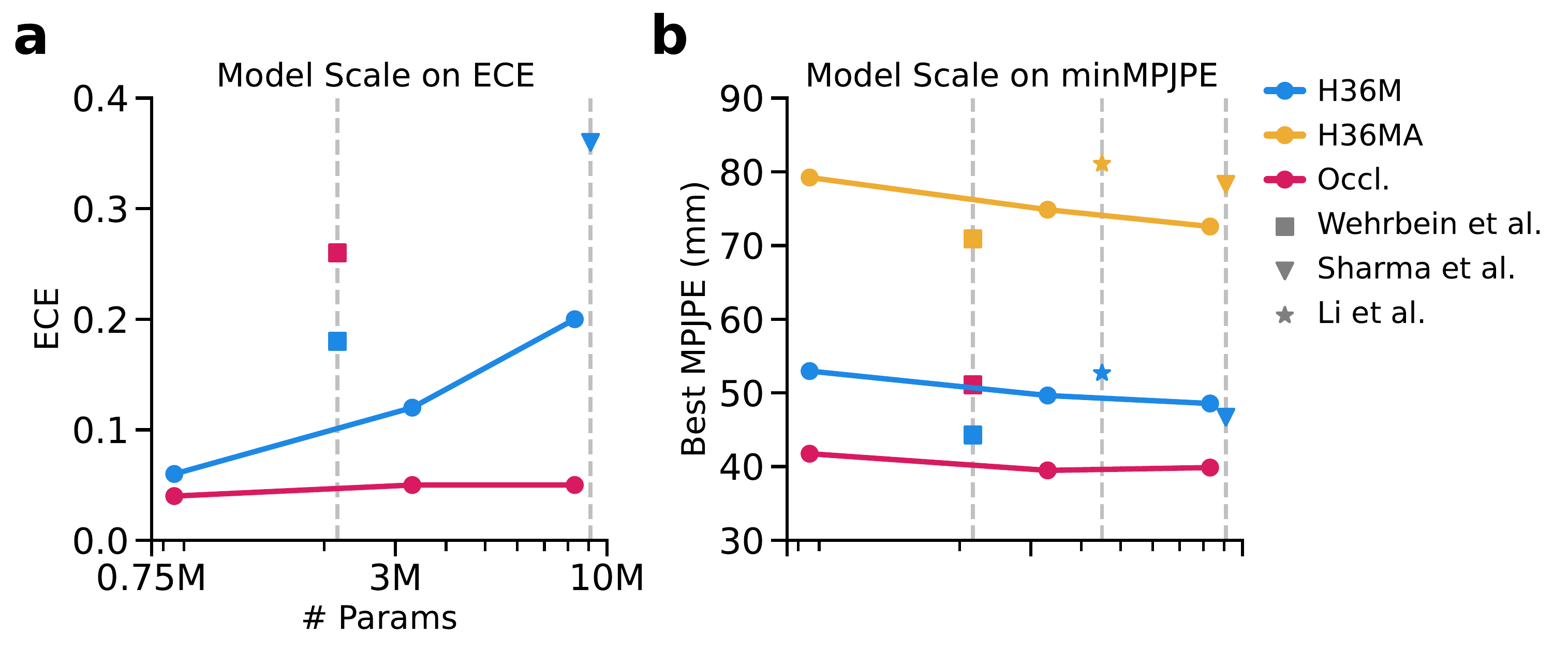}
\caption{$\operatorname{minMPJPE}$ and $\operatorname{ECE}$ across different model sizes. \textbf{a}) shows the effect of model scaling on calibration. \textbf{b}) shows the effect of model scale on accuracy. Performance on both metrics is compared to prior methods at their respective model sizes.
}
\label{fig:model_scaling}
\end{figure}
\subsection{Consequences of model scale}
\label{sec:app_model_scale}
We explore the effect of increasing the number of parameters of the model. We train 3 sizes of models: 1) \textit{small} with 852 546 parameters, 2) \textit{large} with 3 301 546 parameters, and 3) \textit{xlarge} with 8 318 741 parameters. The individual architectures were found by architecture search. We observe that as the size increases the performance of cGNF applied to the lifting task improves decreasing the gap to the state-of-the-art methods.
The performance further improves outperforming the state-of-the-art method on occluded joints.
However, the improvement in performance comes at a cost of calibration (Fig.~\ref{fig:model_scaling}).

\begin{figure}[t!]
\centering
\includegraphics[width=0.6\linewidth]{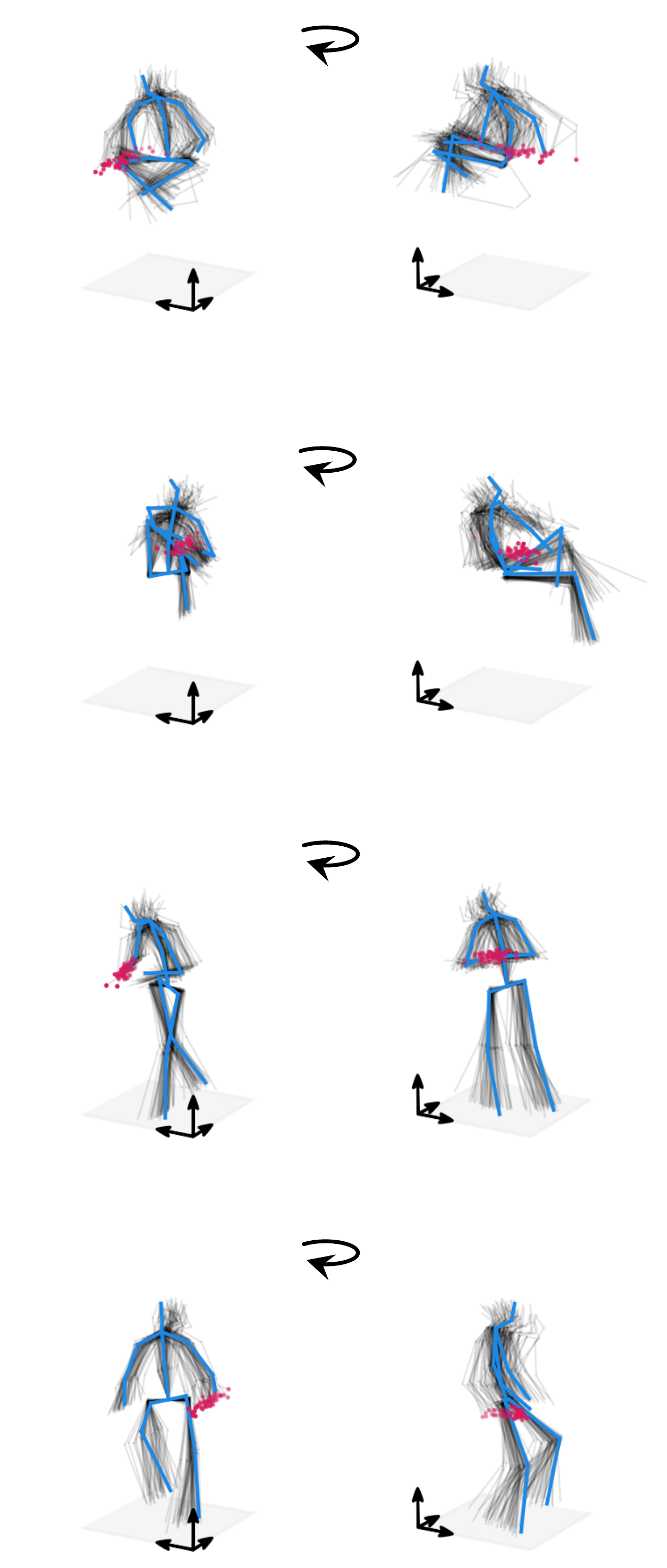}
\caption{Examples of samples from the posterior distribution learned by the cGNF (gray) vs the ground truth pose (blue).
Pink points show the 50 sampled hypotheses for the right wrist positions.
These examples are non-cherry picked and generated for subjects S9 and S11 from the test dataset.}
\label{fig:supp_posterior}
\end{figure}

\begin{figure}[t!]
\centering
\includegraphics[width=\linewidth]{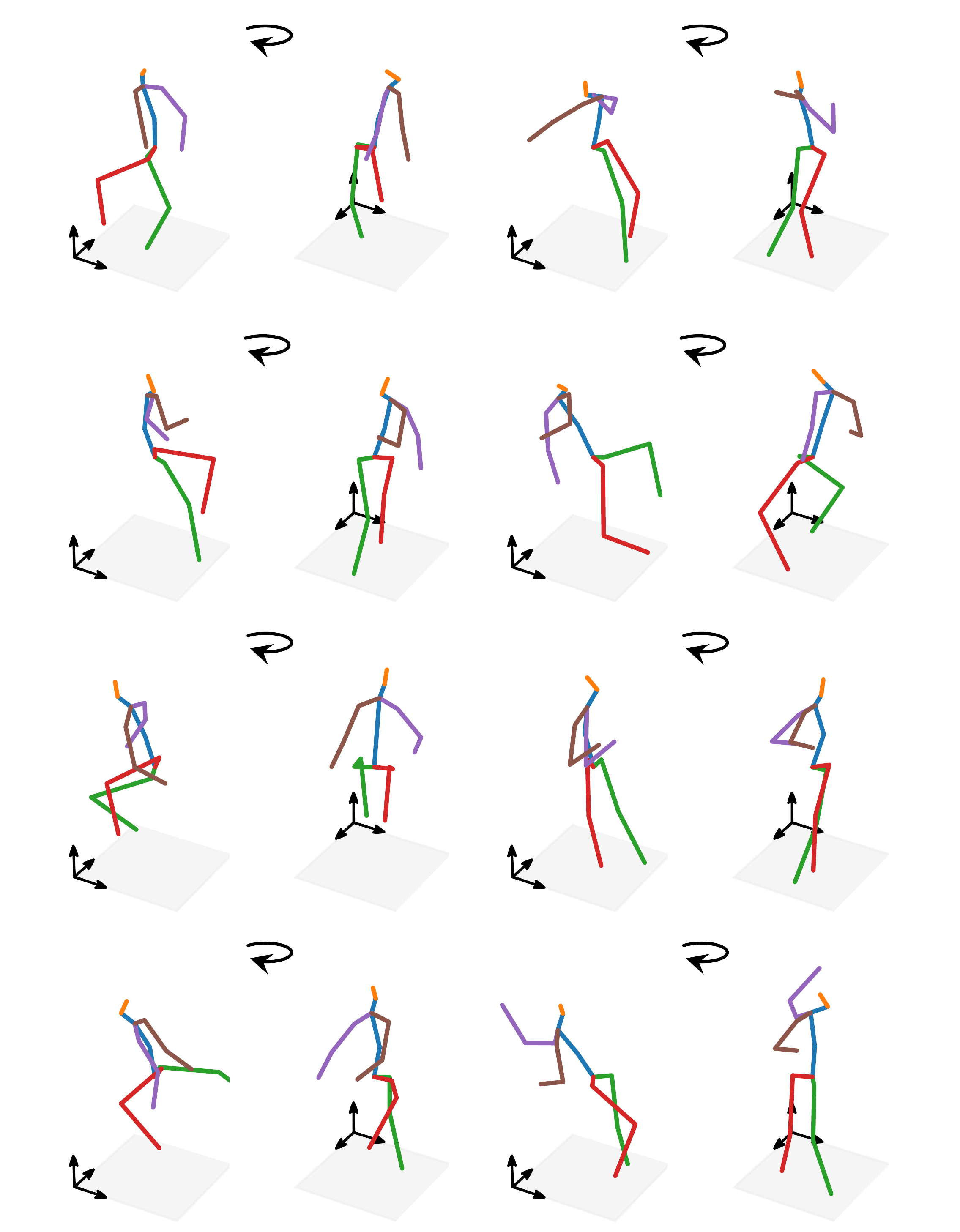}
\caption{Examples of non-cherry picked samples from the prior learned by the cGNF. Each are generated by randomly sampling a latent $\vz \sim \mathcal{N}(\mathbf{0}, \mathbf{I})$ and inverting to the pose space $\mathbf{x} = f^{-1}(\vz, \emptyset)$ without any context $\mathbf{c}$.
For each pose images are shown for two rotations.
}
\label{fig:prior_samples}
\end{figure}

\end{document}